\documentclass[letterpaper]{article} 
\usepackage{aaai2026}  
\usepackage{times}  
\usepackage{helvet}  
\usepackage{courier}  
\usepackage[hyphens]{url}  
\usepackage{graphicx} 
\usepackage{natbib}  
\usepackage{caption} 
\usepackage{algorithm}
\usepackage{algorithmic}

\usepackage{newfloat}
\usepackage{listings}
\DeclareCaptionStyle{ruled}{labelfont=normalfont,labelsep=colon,strut=off} 
\floatstyle{ruled}
\newfloat{listing}{tb}{lst}{}
\floatname{listing}{Listing}
\usepackage[table]{xcolor}
\usepackage{array}
\usepackage{booktabs}
\usepackage{amsmath}
\usepackage[T1]{fontenc}
\usepackage{subcaption}
\usepackage{siunitx}
\usepackage{multirow}
\newcommand{\charbench}{\textsc{CharBench}}

\title{\charbench: Evaluating the Role of Tokenization in Character-Level Tasks}
\author{
  Omri Uzan\textsuperscript{\rm 1},\; Yuval Pinter\textsuperscript{\rm 2}
}
\affiliations{
  \textsuperscript{\rm 1}Stanford University, Stanford, CA, USA\\
  \textsuperscript{\rm 2}Faculty of Computer and Information Science, Ben-Gurion University of the Negev, Beer Sheva, Israel\\
  uzan@stanford.edu\\
}

\usepackage{bibentry}

\begin{document}

\maketitle

\begin{abstract}

Tasks that require character-level reasoning, such as counting or locating characters within words, remain challenging for contemporary language models. A common conjecture is that language models' reliance on subword units, rather than characters, contributes to their struggles with character-level tasks, yet recent studies offer conflicting conclusions about the role of tokenization, leaving its impact unclear. To address this gap, we introduce \charbench{}, a comprehensive benchmark of character-level tasks that is two orders of magnitude larger than existing alternatives.
We evaluate a diverse range of leading open-weight and proprietary models on \charbench{} and find that it presents a significant challenge to modern LLMs, with average accuracies of 43.6\% and 32.3\% on some tasks.
We present an in-depth analysis of how intrinsic properties of words and their segmentations into tokens correspond to model performance. For counting tasks, we find that tokenization properties are weakly correlated with correctness, while the length of the queried word and the actual character count play a more significant part.
In contrast, for tasks requiring intra-word positional understanding, performance is negatively correlated with the length of the token containing the queried character, suggesting that longer tokens obscure information on character position for LLMs.
We encourage future work to build on the benchmark and evaluation methodology introduced here as tools for improving model performance on these tasks.
\end{abstract}

\begin{links}
    \link{Code}{https://github.com/omriuz/CharBench}
    \link{Datasets}{https://huggingface.co/datasets/omriuz/CharBench}
\end{links}

\section{Introduction}

In recent years, the disconnection between large language models (LLMs)' stellar performance on high-level tasks requiring deep understanding of language and their wholly underwhelming ability to perform low-level, even menial tasks at the surface of text analysis, has garnered much attention.
One canonical example of such low-level tasks where LLMs appear to fail spectacularly and unexpectedly is the character counting question, such as 
\textit{How many `r's are there in \texttt{strawberry}?} or \textit{How many `n's are in \texttt{mayonnaise}?}, while such questions are trivially solvable by humans.

While these weaknesses can be mitigated to some extent by prompt engineering or tool usage, for instance by instructing the model to spell or by invoking a code interpreter, investigating why such failures occur can offer deeper insight into fundamental limitations of modern architectures and the inductive biases embedded within them.

\begin{figure*}[t]
\centering
\includegraphics[width=0.6\textwidth]{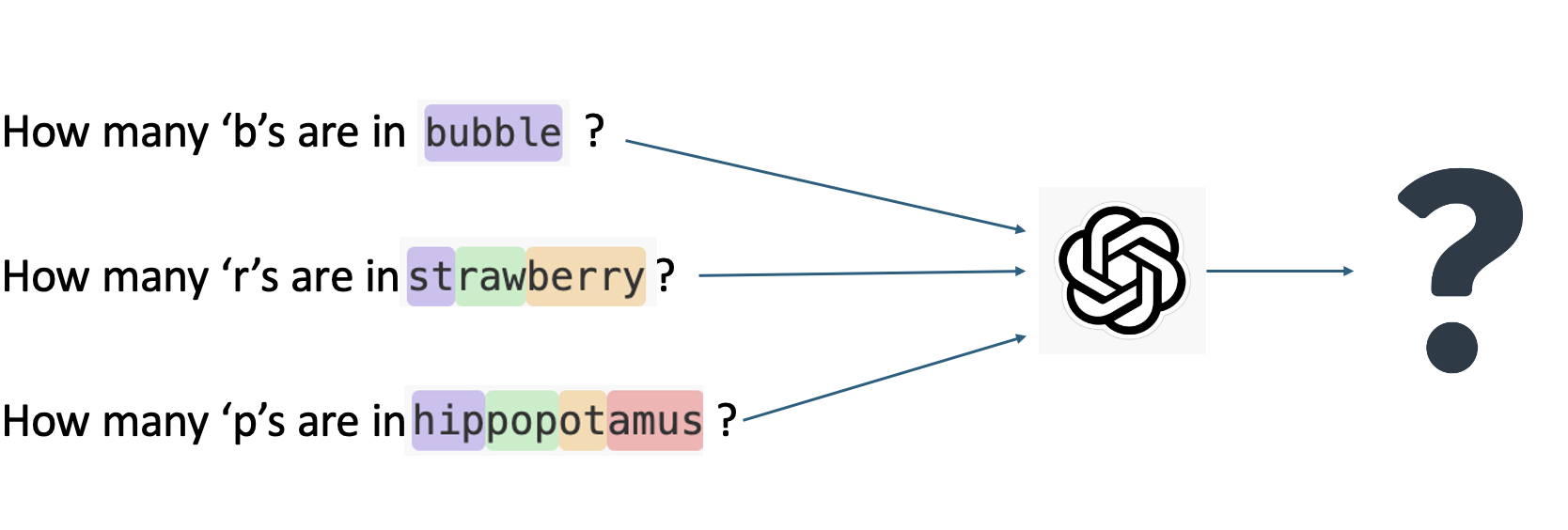}
\caption{Examples of different subword token segmentations for GPT-4o prompts on character counting tasks. Do the properties of words and their segmentation into tokens correlate with model performance?}
\label{fig:diagram}
\end{figure*}

One common conjecture for explaining models' weak performance on character tasks is that language models operate on \emph{subword units} rather than individual characters, and that this mismatch plays a key role in their struggles (as illustrated in Figure~\ref{fig:diagram}).
Prior studies have examined this issue, yet have reached conflicting conclusions about the role of tokenization~\citep{wang2025stringllm, zhang2024countingabilitylargelanguage, shin2024large, xu2025llmgeniusparadoxlinguistic, fu2024largelanguagemodelsllms}, leaving researchers perplexed about its true impact on this type of problem solving.

In order to rigorously investigate this phenomenon, we present \charbench{},
a comprehensive suite of evaluation tasks designed to systematically assess the performance of LLMs on character-level reasoning tasks in English.
\textsc{CharBench} is designed to test LLMs’ ability to reason over characters, with separate tasks for assessing models’ performance on positional understanding (indexing) and occurrence tracking (counting).
We evaluate a diverse set of state-of-the-art language models from both proprietary and open-weight architectures over a range of parameter sizes.
\textsc{CharBench} proves to be a substantial challenge, with an average accuracy of 50.3\% across models.
Positional understanding appears to be particularly difficult, with average accuracies of 43.6\% and 32.4\% in the tasks in this category.

To evaluate the role of tokenization at scale, we build on a line of work investigating intrinsic tokenizer metrics~\citep{galle-2019-investigating,zouhar-etal-2023-tokenization,uzan-etal-2024-greed}. We observe a consistent linear decline in performance as word length increases, echoing previous findings~\citep{fu2024largelanguagemodelsllms} and underscoring a fundamental limitation of current LLMs in processing longer strings.
For counting tasks, we find that performance is strongly correlated with the actual number of character occurrences, while tokenization properties show weaker correlations.
In contrast, for tasks that require locating specific characters, the length of the token containing the target character is the most correlated feature.
We find that accuracy declines as the target token length increases, revealing a tradeoff between token-level compression and the model's ability to reason about character-level structure. Notably, while much tokenization work has focused on properties such as the number of tokens per word and average token length \citep{galle-2019-investigating,goldman-etal-2024-unpacking,schmidt-etal-2024-tokenization,beinborn-pinter-2023-analyzing}, our findings suggest that, for character-level tasks, neither plays a significant role in model performance, and other properties correlate more strongly.

Our findings add a new perspective to the debate in the literature about the role of tokenization in character-level tasks, and complement prior work that frames tokenization as either crucial~\citep{wang2025stringllm,zhang2024countingabilitylargelanguage,shin2024large} or insignificant~\citep{xu2025llmgeniusparadoxlinguistic,fu2024largelanguagemodelsllms}.
We show that its relationship with surface form is substantial for certain tasks, and less meaningful for others.
We hope this benchmark and evaluation framework serves as a foundation for future work aimed at better understanding and improving model performance on these tasks.

\section{Related Work}

\paragraph{Intrinsic Measures of Tokenization.} 
Prior work has investigated intrinsic properties of tokenization, primarily focusing on tokenizer behavior across word sequences.
\citet{galle-2019-investigating} argue that the effectiveness of the byte-pair encoding algorithm~\citep[BPE;][]{sennrich-etal-2016-neural} in translation tasks stems from its ability to represent word sequences with fewer symbols.
\citet{zouhar-etal-2023-tokenization} show that Rényi entropy over token distributions across word sequences correlates more strongly with downstream performance than token count alone.
However, subsequent studies provide counterexamples where optimizing for token count or Rényi entropy does not lead to improved downstream performance~\citep{cognetta-etal-2024-two,schmidt-etal-2024-tokenization}.
Alongside information theory-based measures, \citet{uzan-etal-2024-greed} also integrate intrinsic metrics from cognitive science~\citep{beinborn-pinter-2023-analyzing} and morphology~\citep{gow-smith-etal-2022-improving} into a unified benchmark.
Nonetheless, these approaches predominantly assess tokenizer performance at the word-sequence level, overlooking intra-word dynamics that are informative for character-level tasks.

\paragraph{Benchmarking Character-Level Performance in LLMs.} 
While other benchmarks for character-level tasks have been proposed, none that we know of are well-suited for large-scale statistical evaluation of tokenizer properties in relation to downstream performance.
\citet{efrat-etal-2023-lmentry} introduce \textsc{LMentry}, which includes some character-level tasks, but these primarily target the first or last character in words, limiting their ability to generalize to broader tokenization behavior.
\citet{chai-etal-2024-tokenization} present a benchmark with several token-structure probing tasks, containing roughly 300 words, two orders of magnitude below our \charbench.
\textsc{CUTE}~\citep{edman-etal-2024-cute} provides a broader set of challenging tasks, yet it is intentionally limited to mostly frequent, single-token words. 
\citet{wang2025stringllm} take a similar approach to ours by automatically generating a large-scale collection of verifiable question-answer pairs involving natural language operations;
however, their benchmark primarily targets word-sequence performance rather than character-level or subword-level phenomena.

\paragraph{Tokens and Characters.}
While the extent to which subword tokens encode character-level information has been thoroughly studied at the embedding level, methods for analyzing this relationship in downstream generation remain limited.
\citet{itzhak-levy-2022-models} show that subword-based pretrained models can recover correct spellings, suggesting that token embeddings retain some fine-grained character-level information.
\citet{kaushal-mahowald-2022-tokens} further support this claim by probing token embeddings for individual characters, attributing this capacity to variation in subword tokenization across related strings during pretraining.
More recent work~\citep{xu2025llmgeniusparadoxlinguistic,zhang2024countingabilitylargelanguage} explores the link between subword tokenization and character-level performance but arrives at conflicting conclusions regarding its significance.
A key limitation shared by these approaches is their reliance on explicit perturbations, adding special characters to induce character-level segmentation, which introduces noise, diverges from the model's training distribution, inflates input length, and ultimately hinders direct comparability.
\\
\definecolor{lightgray}{gray}{0.95}
\definecolor{lightblue}{RGB}{220,230,241}
\definecolor{lightgreen}{RGB}{220,241,220}

{
\begin{table*}[t]
\centering
\small
\rowcolors{2}{lightgray}{white}
\begin{tabular}{>{\centering\arraybackslash}p{3cm} >{\centering\arraybackslash}p{7cm} >{\centering\arraybackslash}p{1.5cm}}
\toprule
\rowcolor{gray!20}
\textbf{Task} & \textbf{Example} & \textbf{Label} \\
\midrule
Character Frequency Count & How many times does the character \texttt{`s'} appear in the string \texttt{``mississippi''}? & 4 \\
Unique Character Count & How many unique characters appear in the string \texttt{`balloon'}?  & 5 \\
Find First Occurrence & What is the index of the first occurrence of the character \texttt{`e'} in the string \texttt{`cheese'}? Start counting from 0. & 2 \\
Find Last Occurrence & What is the index of the last occurrence of the character \texttt{`n'} in the string \texttt{`cinnamon'}? Start counting from 0. & 7 \\
\bottomrule
\end{tabular}
\caption{Example questions from \charbench. One question is shown per task. Words and characters are interchangeable within the prompt template.}
\label{fig:benchmark_examples}
\end{table*}
}

We use intrinsic metrics of tokenization that emphasize intra-word dynamics, perform a large-scale statistical evaluation with control for confounding factors, and assess tokenizers' role from a purely statistical perspective without perturbing input strings, allowing us to draw more statistically robust conclusions.

\section{\charbench}

The core insight in the construction of \charbench{} is that many character-level reasoning tasks are inherently deterministic, making them well-suited for large-scale, controlled evaluations.
While these tasks are presented in natural language, for example, \textit{``How many \texttt{r}'s are in `strawberry'?''}, they can be directly translated into more simple computational operations, such as \texttt{count(`r', `strawberry')}.
By exploiting this deterministic and verifiable structure, \charbench{} enables the scalable construction of question-answer pairs while controlling for confounding factors such as word length, frequency, and token count.

To construct \charbench{}, we developed a set of modular task templates that represent a variety of character-level reasoning scenarios.
These templates can be instantiated with arbitrary words and characters, as illustrated in Figure~\ref{fig:benchmark_examples}.
We focused on tasks that are verifiable and can be mapped directly to simple Python functions, such as \texttt{count()} or \texttt{index()}.
To populate the benchmark, we uniformly sampled 175,000 strings from the MiniPile dataset~\citep{kaddour2023minipilechallengedataefficientlanguage}, ensuring balanced coverage across lengths from 4 to 10 characters.
This stratified sampling strategy mitigates potential biases related to word length and frequency.

\charbench{} evaluates two types of character-level information within words: \textit{occurrence information}, which captures the presence of specific characters, and \textit{positional information}, which reflects the locations of characters within a word (e.g., which character appears at a given index).
Previous work on character knowledge in token embeddings has primarily examined character occurrence~\citep{kaushal-mahowald-2022-tokens} and relative spelling patterns~\citep{itzhak-levy-2022-models}, without explicitly evaluating whether models encode information about absolute character positions (while spelling patterns capture relative order, they do not require awareness of absolute positions within a word).

To capture both quantitative and positional aspects of word-level character knowledge, \charbench{} includes two task categories: \textit{counting} and \textit{indexing}.
Counting tasks assess an LLM’s ability to determine character frequencies within strings, simple for humans on short words but often challenging for language models.
These include counting the occurrences of a specific character and identifying the set of unique characters in a word.
Indexing tasks evaluate positional understanding by requiring the model to locate the position (index) of either the first or last occurrence of a specified character within a string. While it is relatively simple to construct more tasks, we focus on two tasks per aspect to enable simpler qualitative analysis of models' performance.



\section{Experiments}

\begin{table*}[t]
\centering
\small
\begin{tabular}{lccccc}
\toprule
\textbf{Model} 
& \textbf{Overall} 
& \textbf{count\_char} 
& \textbf{count\_unique} 
& \textbf{find\_first} 
& \textbf{find\_last} \\
\midrule
\textbf{GPT-4o}
 & \textbf{70.73\%} 
 & \textbf{89.45\%} 
 & \textbf{65.51\%} 
 & \textbf{64.65\%} 
 & \textbf{63.14\%} \\
\textbf{GPT-4o-mini}
 & 49.99\% 
 & 82.54\% 
 & 47.32\% 
 & 42.89\% 
 & 27.07\% \\
\textbf{GPT-3.5-turbo}
 & 45.22\% 
 & 79.11\% 
 & 52.08\% 
 & 28.96\% 
 & 20.72\% \\
\textbf{DeepSeek-V3}
 & 57.18\% 
 & 74.52\% 
 & 57.31\% 
 & 57.85\% 
 & 38.72\% \\
\textbf{Llama-3.3-70B}
 & 39.98\% 
 & 61.90\% 
 & 43.07\% 
 & 34.11\% 
 & 20.74\% \\
\textbf{Meta-Llama-3.1-405B}
 & 55.49\% 
 & 81.21\% 
 & 44.83\% 
 & 54.24\% 
 & 41.01\% \\
\textbf{Mistral-7B}
 & 34.72\% 
 & 71.65\% 
 & 30.63\% 
 & 20.76\% 
 & 15.89\% \\
\midrule
\textbf{Average}
 & 50.33\% 
 & 77.34\% 
 & 47.96\% 
 & 43.64\% 
 & 32.37\% \\
\bottomrule
\end{tabular}
\caption{Model results across the different tasks in \charbench. Tasks are identified by their prefixes and are ordered consistently with Table~\ref{fig:benchmark_examples}.}
\label{tab:benchmark_results}
\end{table*}

We evaluate a diverse set of state-of-the-art language models on \charbench.
We include both proprietary and open-weight architectures, selected based on the availability of \emph{open tokenizer access}, which is critical for our analysis of the interaction between a model and the tokenizer it has access to.
All selected models use the widely-adopted byte-pair encoding (BPE) tokenization scheme. 

\paragraph{Open-weight Models.} 
We evaluate several prominent open-weight models: DeepSeek-V3 \citep{deepseekai2025deepseekv3technicalreport}, Llama-3.3-70B, Llama-3.1-405B \citep{grattafiori2024llama3herdmodels}, and Mistral-7B \citep{jiang2023mistral7b}.
These were selected to represent a range of model sizes and architectural designs.
All models are accessed via the \texttt{Together.AI} API,\footnote{\url{https://www.together.ai/}} ensuring a consistent inference environment and enabling evaluation of large-scale models that would otherwise require substantial local compute resources.

\paragraph{Proprietary Models.} 
We include OpenAI's GPT-4o~\citep{openai2024gpt4ocard}, GPT-4o-mini, and GPT-3.5-turbo in our evaluation, as these models provide unrestricted access to their tokenization mechanism via the \texttt{tiktoken} library.\footnote{\url{https://github.com/openai/tiktoken}}
Other state-of-the-art proprietary models were excluded due to restrictions on tokenizer access, which prevents an analysis of tokenization effects.

\paragraph{Evaluation Protocol.} 
Model performance is quantified using accuracy, defined as the proportion of predictions matching the gold-standard label (exact match).

\paragraph{Parameter Settings.} 
To ensure fair comparison and reproducibility, we standardize all evaluation settings across models.
We set the temperature to 0 to eliminate sampling variability and produce deterministic outputs;
we use uniform system and user prompt templates for all models;
we maintain consistent API usage and parameter settings across both open-weight and proprietary models.

\subsection{Results}

\paragraph{Model Performance.} \charbench{} presents a substantial challenge for modern language models, with an average accuracy of 50.33\% across all evaluated models.
GPT-4o achieves the highest performance by a significant margin, attaining an average accuracy of 70.73\% and outperforming all other models across tasks.
Among open-weight models, DeepSeek-V3 achieves the strongest overall performance, outperforming others across most benchmark tasks.
Llama-3.1-405B follows closely, and surpasses DeepSeek-V3 on the \texttt{count\_char} and \texttt{find\_last} tasks.
Notably, Mistral-7B trails Llama-3.3-70B by only 5 points on average, despite being 10 times smaller in parameter count, and even surpasses it by 10 points on the \texttt{count\_char} task.
We also find that GPT-4o-mini exhibits a substantial drop in character-level performance compared to GPT-4o, trailing by an average of 20 points.
Complete results are provided in Table~\ref{tab:benchmark_results}.

\paragraph{Task Performance.} Tasks that require absolute positional understanding over characters, namely \texttt{find\_first} and \texttt{find\_last}, prove significantly more challenging for models than counting tasks. While many prior works have focused on character-level understanding through character occurrence, 
our results indicate that models struggle more with tracking character positions.
We consistently observe a performance gap between \texttt{find\_first} and \texttt{find\_last}, with the former being solved more reliably.
This gap also varies significantly across models, showing an almost 20\% difference for DeepSeek-V3, compared to just 0.5\% for GPT-4o.
Notably, the canonical question \emph{``How many `r's are there in \texttt{strawberry}?''} (\texttt{count\_char}), although still far from solved, emerges as the easiest task for models on \charbench, by a substantial margin.

\section{Analysis}

\sisetup{
  table-number-alignment = center,
  table-format = -1.3,
  detect-all
}
\begin{table*}[t]
\centering
\footnotesize
\begin{tabular}{
  l
  l
  S[table-format=-1.3]
  S[table-format=-1.3]
  S[table-format=-1.3]
  S[table-format=-1.3]
  S[table-format=-1.3]
  S[table-format=-1.3]
}
\toprule
task & model & {WL} & {GT} & {GT/WL} & {NT} & {CR} & {TTL} \\
\midrule
\multirow{8}{*}{\centering Find First Occurrence}
  & Llama-405B    & -0.212 & \textbf{-0.325} & -0.277 &  0.004 & -0.191 & -0.317 \\
  & Llama-70B     & -0.257 & -0.237 & -0.167 & -0.010 & -0.208 & \textbf{-0.275} \\
  & DeepSeek-V3   & -0.184 & \textbf{-0.542} & -0.511 & -0.025 & -0.152 & -0.291 \\
  & Mistral-7B    & -0.134 & \textbf{-0.259} & -0.247 & -0.022 & -0.095 & -0.048 \\
  & GPT-4o-mini   & -0.147 & -0.483 & \textbf{-0.497} & -0.006 & -0.139 & -0.242 \\
  & GPT-3.5       & \textbf{-0.223} & -0.126 & -0.041 & -0.003 & -0.175 & -0.194 \\
    \cmidrule(lr){2-8}
  & GPT-4o (best-performing) & -0.136 & -0.170 & -0.128 &  0.059 & -0.158 & \textbf{-0.245} \\
  & Average       & -0.185 & \textbf{-0.306} & -0.267 &  0.000 & -0.160 & -0.230 \\
\midrule
\multirow{8}{*}{\centering Find Last Occurrence}
  & Llama-405B    & -0.247 & -0.092 & -0.017 & -0.028 & -0.197 & \textbf{-0.267} \\
  & Llama-70B     & -0.168 & 0.279 & \textbf{0.433} & -0.050 & -0.109 & -0.080 \\
  & DeepSeek-V3   & \textbf{-0.267} & -0.126 & -0.042 & -0.041 & -0.205 & -0.242 \\
  & Mistral-7B    & \textbf{-0.099} &  0.009 & 0.086 & -0.015 & -0.067 & -0.019 \\
  & GPT-4o-mini   & -0.201 & 0.258 & \textbf{0.420} & -0.067 & -0.127 & -0.161 \\
  & GPT-3.5       & -0.135 & 0.295 & \textbf{0.468} & -0.094 & -0.048 & -0.010 \\
    \cmidrule(lr){2-8}
  & GPT-4o (best-performing) & -0.180 & -0.126 & -0.066 &  0.020 & -0.185 & \textbf{-0.259} \\
  & Average       & \textbf{-0.185} & 0.071 & 0.183 & -0.039 & -0.134 & -0.148 \\
\midrule
\multirow{8}{*}{\centering Character Frequency Count}
  & Llama-405B    & -0.242 & \textbf{-0.478} & -0.244 & -0.035 & -0.183 & {--} \\
  & Llama-70B     & \textbf{-0.364} & -0.011 & 0.212 & -0.099 & -0.219 & {--} \\
  & DeepSeek-V3   & -0.125 & \textbf{-0.205} & -0.091 & -0.025 & -0.088 & {--} \\
  & Mistral-7B    & -0.240 & \textbf{-0.531} & -0.319 & -0.106 & -0.099 & {--} \\
  & GPT-4o-mini   & -0.194 & \textbf{-0.359} & -0.175 & -0.015 & -0.175 & {--} \\
  & GPT-3.5       & -0.201 & \textbf{-0.228} & -0.090 & -0.067 & -0.106 & {--} \\
    \cmidrule(lr){2-8}
  & GPT-4o (best-performing) & -0.099 & \textbf{ -0.340} & -0.243 &  0.009 & -0.083 & {--} \\
  & Average       & -0.209 & \textbf{-0.307} & -0.136 & -0.048 & -0.136 & {--} \\
\midrule
\multirow{8}{*}{\centering Unique Character Count}
  & Llama-405B    & -0.433 & -0.122 & \textbf{0.669} & -0.189 & -0.222 & {--} \\
  & Llama-70B     & -0.488 & -0.198 & \textbf{0.642} & -0.193 & -0.269 & {--} \\
  & DeepSeek-V3   & \textbf{-0.317} & -0.271 & 0.114 & -0.106 & -0.195 & {--} \\
  & Mistral-7B    & -0.262 & \textbf{-0.364} & -0.136 & -0.048 & -0.178 & {--} \\
  & GPT-4o-mini   & -0.517 & -0.213 & \textbf{0.663} & -0.243 & -0.257 & {--} \\
  & GPT-3.5       & -0.364 & \textbf{-0.366} & 0.102 & -0.116 & -0.230 & {--} \\
    \cmidrule(lr){2-8}
  & GPT-4o (best-performing) & -0.301 & -0.099 & \textbf{0.462} & -0.108 & -0.183 & {--} \\
  & Average       & \textbf{-0.383} & -0.233 & 0.359 & -0.143 & -0.219 & {--} \\
\bottomrule
\end{tabular}
\caption{Correlation coefficients between correctness and input properties. Abbreviations: WL = word length; GT = gold truth; GT/WL = gold truth divided by word length; NT = number of tokens; CR = compression rate; TTL = target token length. Dashes (--) denote not applicable values (e.g., TTL for counting tasks). Bold indicates the strongest absolute correlation per row.}
\label{tab:task_model_correlations}
\end{table*}

\begin{figure}[t]
  \centering
  \includegraphics[width=\linewidth]{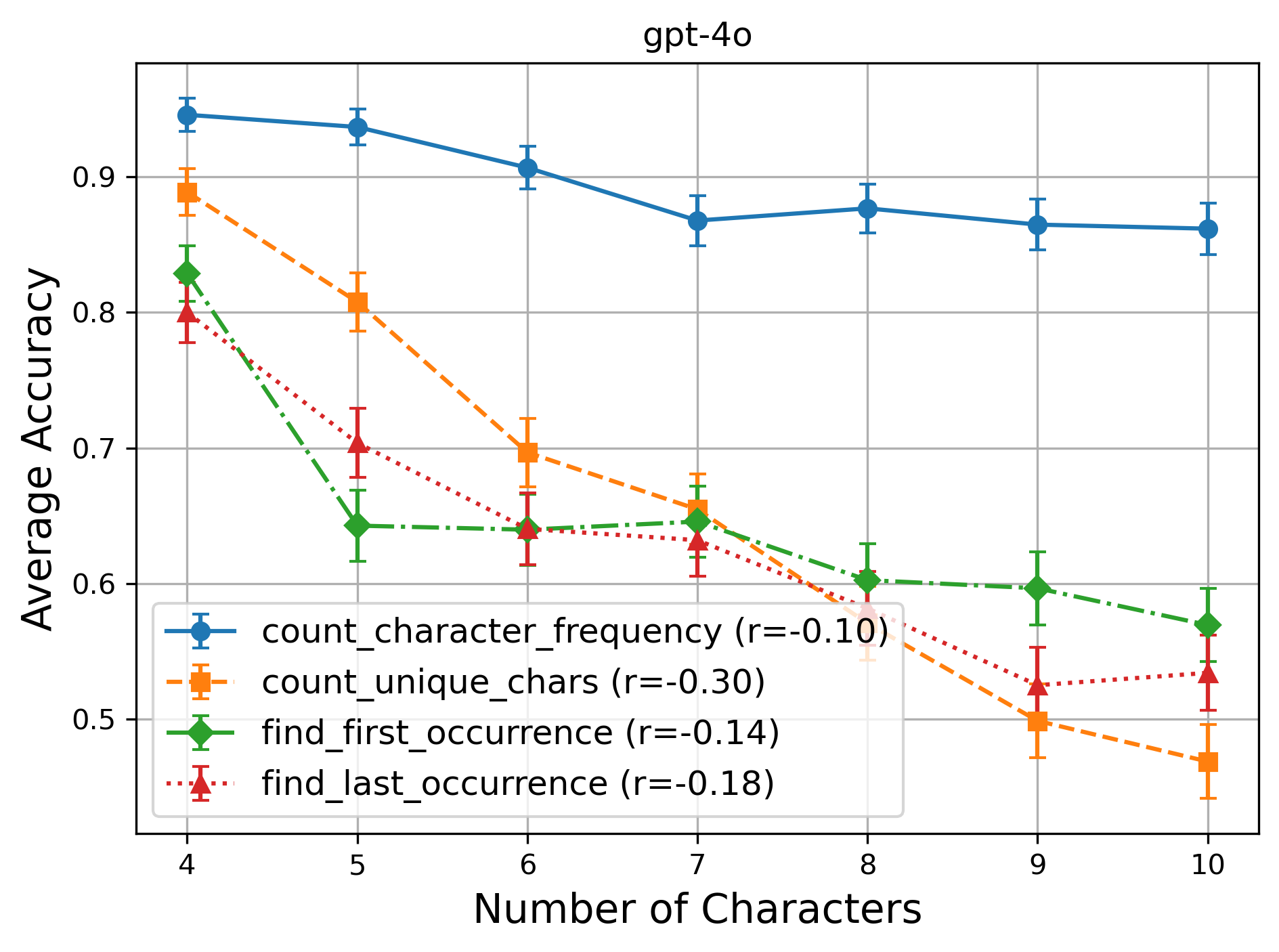}
  \caption{GPT-4o's accuracy on all four tasks as a function of word length.}
  \label{fig:word_length}
\end{figure}

We examine how model performance on \charbench{} correlates with intrinsic properties of the queried word and its tokenization, as well as task-specific attributes such as the gold truth and the queried character when applicable.
We treat each prediction as a binary indicator (1 = correct, 0 = incorrect); we quantify its association with intrinsic properties using Pearson's Correlation.
We examine the following properties:
\begin{figure*}[t]
  \centering
  \begin{subfigure}[t]{0.48\textwidth}
    \centering
    \includegraphics[width=\linewidth]{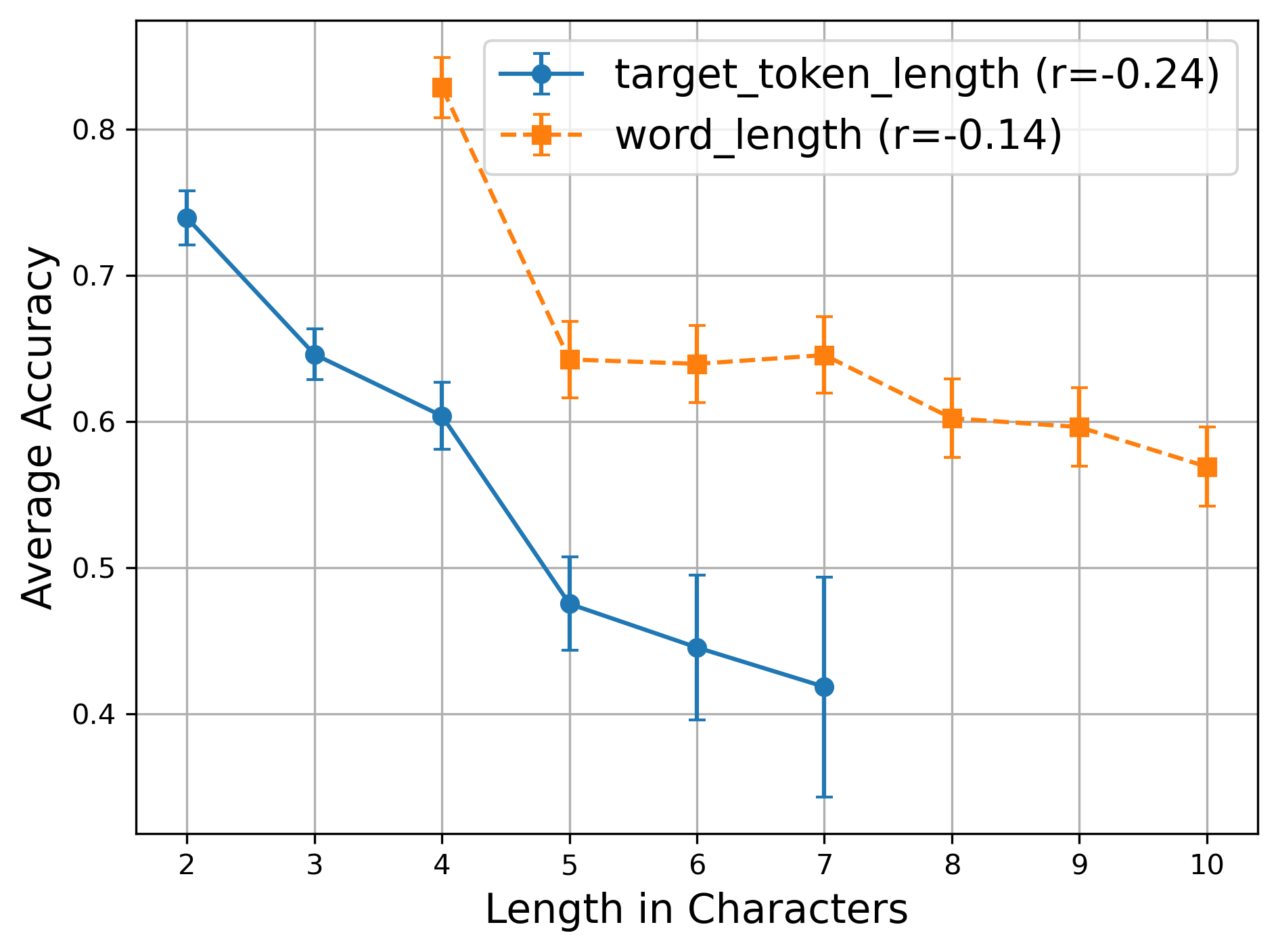}
    \caption{Find First Occurrence}
    \label{fig:first_occ}
  \end{subfigure}
  \hfill
  \begin{subfigure}[t]{0.48\textwidth}
    \centering
    \includegraphics[width=\linewidth]{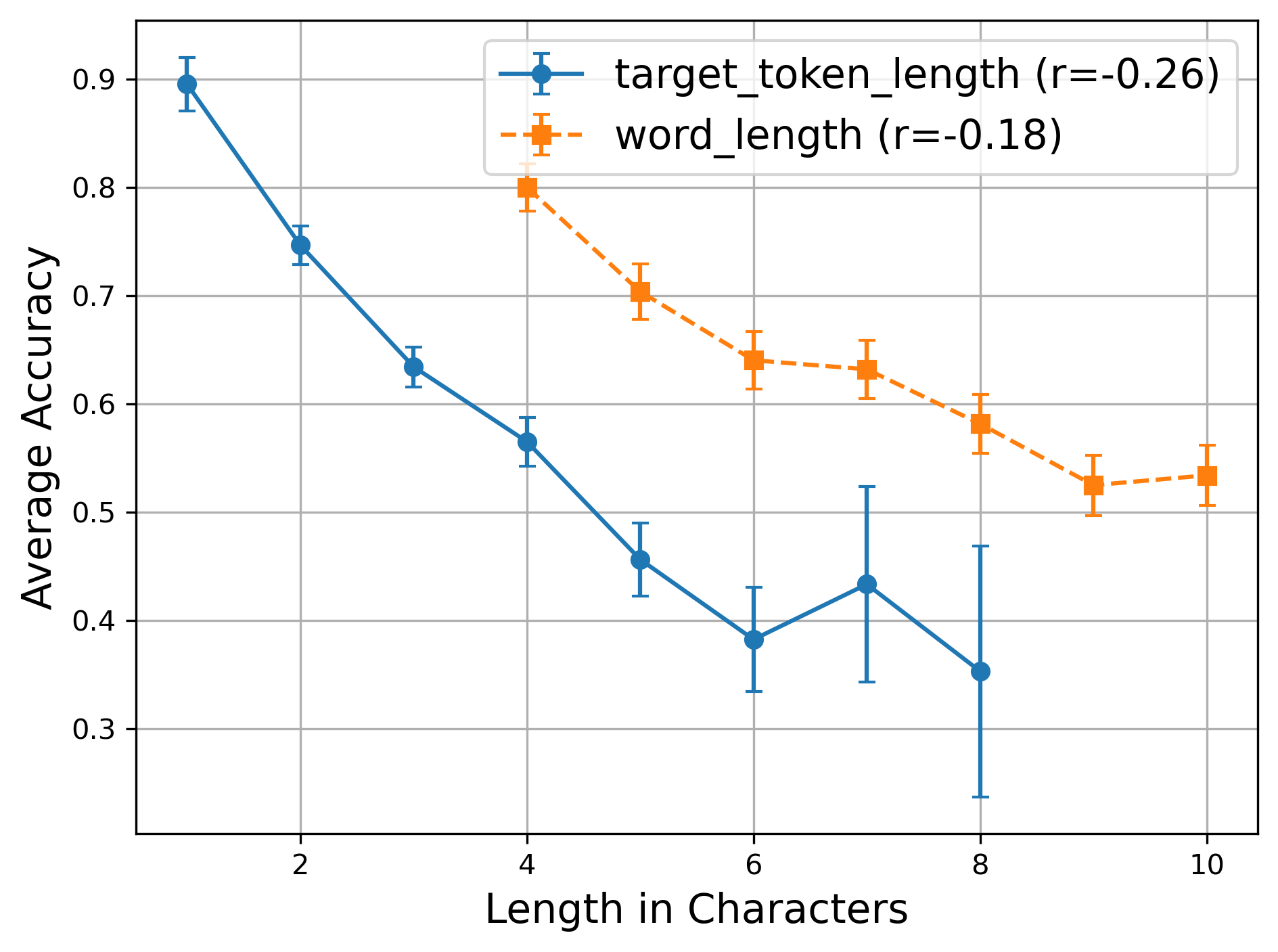}
    \caption{Find Last Occurrence}
    \label{fig:last_occ}
  \end{subfigure}
\caption{Accuracy as a function of word length, plotted separately for each task.
Points show the mean accuracy for all items of a given length. Error bars
indicate standard error of the binomial proportion for that bucket.}
  \label{fig:accuracy_by_length_task}
\end{figure*}

\paragraph{Word Length (WL),} the number of characters in the word.
\paragraph{Gold Truth (GT),} the correct answer for the given question. In indexing tasks, this is the index of the queried character. In counting tasks, this is the true count of the target character(s).
\paragraph{Gold Truth Divided by Word Length (GT/WL),} a normalized variant of GT.
In indexing tasks, it reflects the relative position of the queried character within the word (i.e.,~small values for the start of the word and high values for the ending).
In the \texttt{Character Frequency Count} task, it indicates the proportion of the word occupied by the target character.
In the \texttt{Unique Character Count} task, it captures the proportion of unique characters in the word. From each word's segmentation into tokens, we derive the following tokenization-level properties:

\paragraph{Number of Tokens (NT),} the total number of tokens the word is split into via tokenization.

\paragraph{Compression Ratio (CR),} the ratio of character length to token count, indicating the average number of characters per token.

\paragraph{Target Token Length (TTL),} The character length of the token that contains the queried character.
This is defined only for indexing tasks that reference a specific character position.

Table~\ref{tab:task_model_correlations} reports, for each \charbench{} task and model, the correlation between model performance and each intrinsic property. For aggregation, it presents two complementary indicators: the \emph{average correlation} across all models, which reflects general trends, and the \emph{correlation for the best-performing model} on the benchmark, GPT-4o in all tasks, which highlights how properties relate to performance in a strong model.
To better illustrate the relationship between performance and these intrinsic properties, we analyze the average performance across property values for the best-performing model on \charbench{}, GPT-4o.
In figures below, we plot the mean accuracy for each property value, with error bars showing standard error computed from the bucket.

\subsection{Word Length Consistently Correlates Negatively with Performance }

\textbf{Word length} exhibits a consistent correlation with performance across models and tasks, echoing findings by \citet{fu2024largelanguagemodelsllms}.

Figure~\ref{fig:word_length} illustrates this relationship for GPT-4o, showing that longer words are consistently associated with lower average prediction accuracy. As shown in Table~\ref{tab:task_model_correlations}, the \texttt{Unique Character Count} task displays the strongest negative correlation with word length, with an average correlation of $-0.383$ across models.

Other tasks exhibit weaker correlations, generally ranging between $-0.18$ and $-0.21$. This supports the notion that tasks requiring holistic processing of the word, such as \texttt{Unique Character Count}, are more sensitive to word length than tasks focused on localized character-level operations.

\subsection{The Gold Truth Effects}

\begin{figure}[t]
  \centering
  \includegraphics[width=\linewidth]{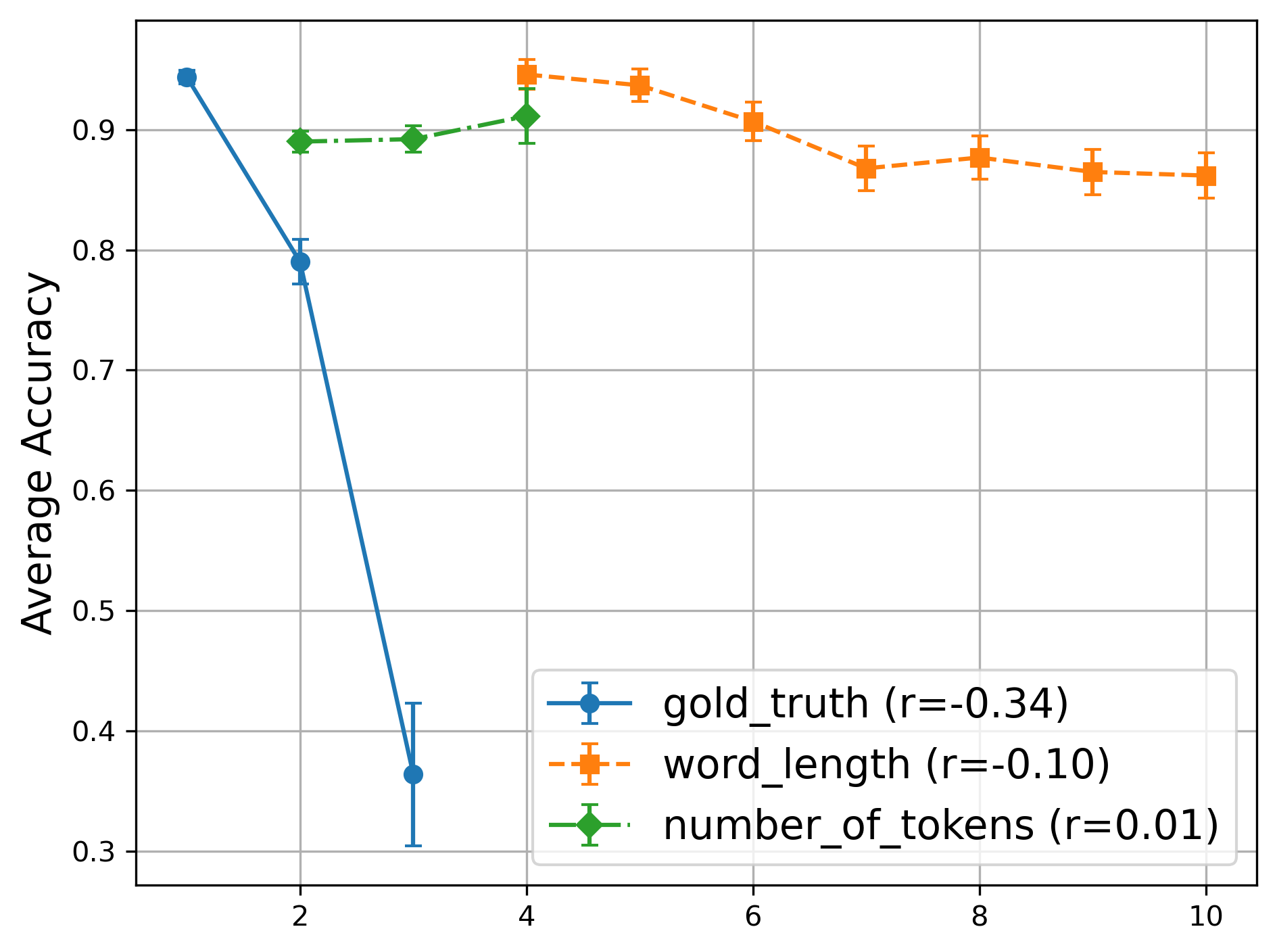}
  \caption{Accuracy as a function of word length, number of tokens, and character spread on the \textit{count character frequency} task.}
  \label{fig:accuracy_character_freq}
\end{figure}

Figure~\ref{fig:accuracy_character_freq} compares the relationship between GPT-4o's performance on the \texttt{Character Frequency Count} task and three intrinsic properties: word length, number of tokens, and the gold truth.
We find that performance on this task for GPT-4o is uncorrelated with the total number of tokens and only weakly negatively correlated with word length ($-0.10$). In contrast, it shows a notably stronger negative correlation with the \textbf{Gold Truth (GT)} metric, which measures how many such characters are present in the word. This trend is consistent across all models except Llama-70B, as reflected in Table~\ref{tab:task_model_correlations}.

For the \texttt{Unique Character Count} task, we find that the gold truth normalized by the length of the word \textbf{(GT/WL)} is positively correlated with performance, with correlation values of over 0.64 for several models. This suggests that as more characters in the word are unique, so it is easier for the model to predict the answer.

For indexing tasks, the \textbf{GT} and \textbf{GT/WL} are also correlated with the average performance, yet mostly so for the weaker models.
The best performing model (GPT-4o) is less correlated with the gold truth variants.


\begin{figure*}[t]
  \centering
  \begin{subfigure}[t]{0.45\textwidth}
    \centering
    \includegraphics[width=\linewidth]{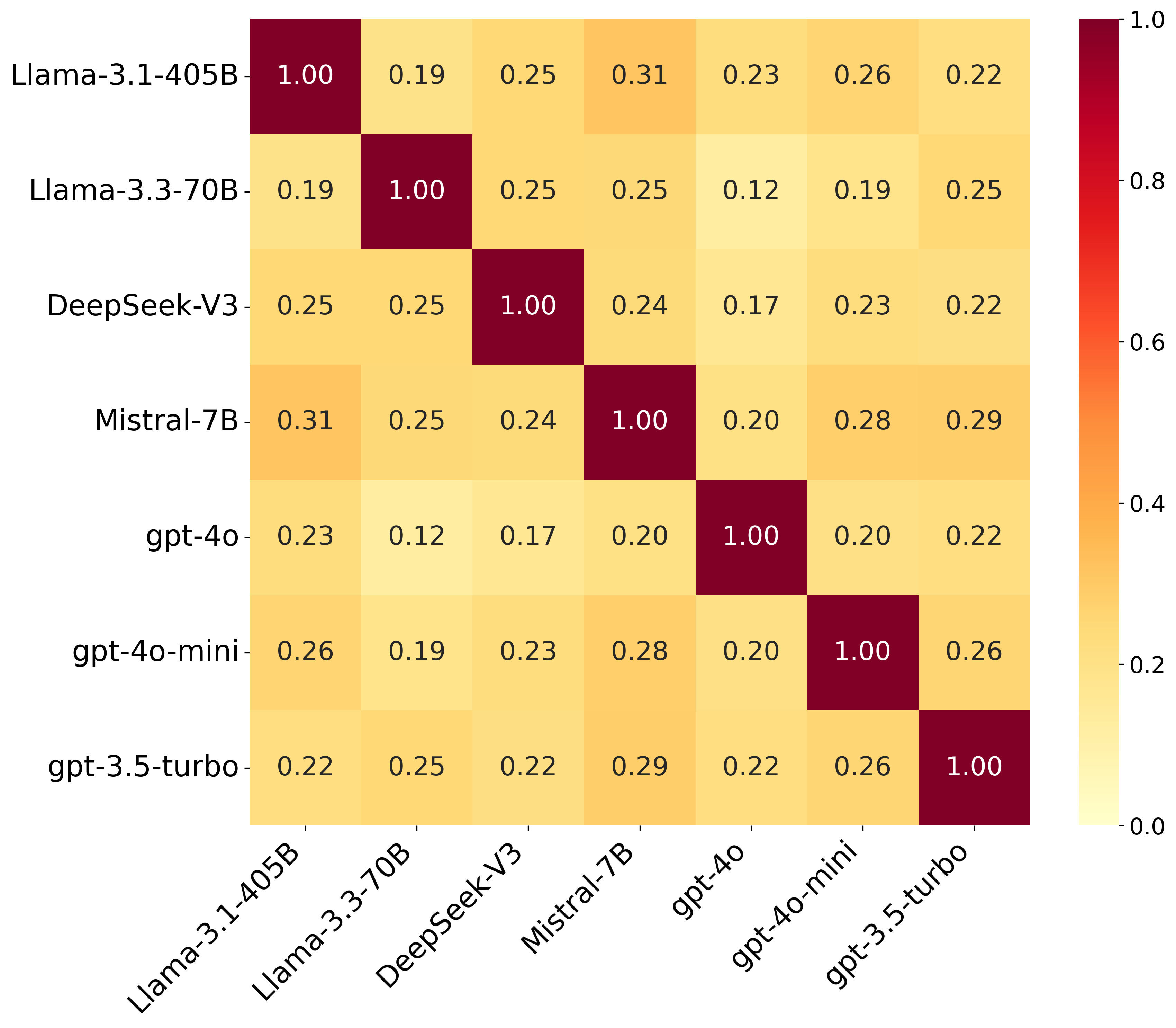}
    \caption{Count Character Frequency}
    \label{fig:heatmap1}
  \end{subfigure}
  \hfill
  \begin{subfigure}[t]{0.45\textwidth}
    \centering
    \includegraphics[width=\linewidth]{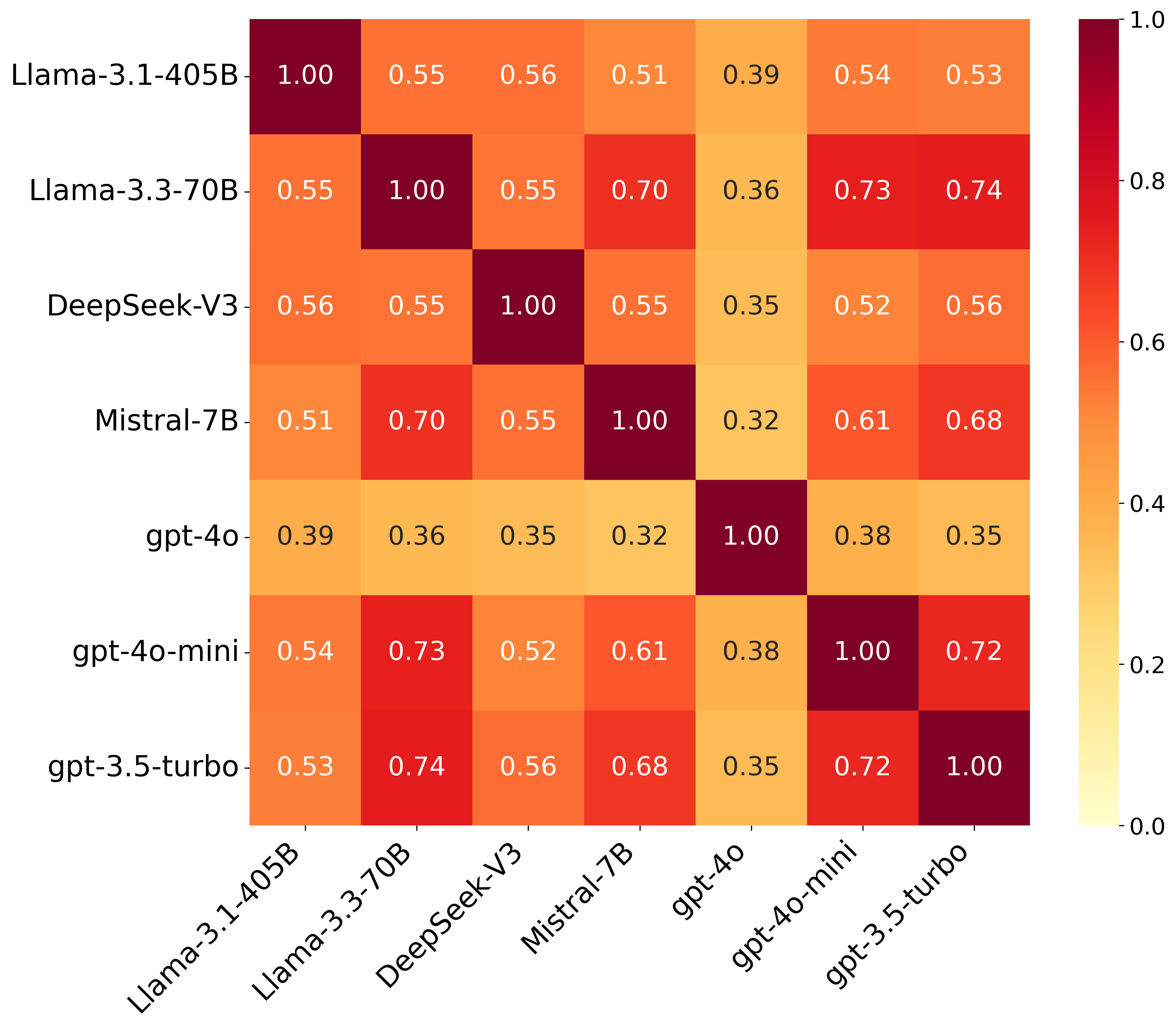}
    \caption{Find Last Occurrence}
    \label{fig:heatmap2}
  \end{subfigure}
\caption{Error overlap between models, measured as the intersection-over-union of incorrectly answered questions for each model pair.}

  \label{fig:error_overlap_heatmap}
\end{figure*}

\subsection{Longer Tokens Obscure Character Position Information for LLMs}

We analyze the effect of token length on model performance on indexing tasks, which require identifying a specific character in the input.
These tasks inherently depend on correctly locating the token that contains the target character.
For the best-performing model, GPT-4o, we find that the length of this \emph{target token} is the most strongly correlated feature with accuracy.

Figure~\ref{fig:accuracy_by_length_task} presents the average performance as a function of both word length and target token length.
We can see that performance declines sharply as the target token becomes longer.
This finding is notable because tokenizers are typically optimized to compress character sequences into fewer tokens to improve computational efficiency. Our analysis suggest a tradeoff between token-level compression and the model’s ability to reason about character-level positions within words.

\subsection{Number of Tokens and Compression Ratio Are Weakly Correlated with Performance}

Across tasks and models, Table~\ref{tab:task_model_correlations} shows that both the number of tokens a word is split into and the compression ratio (i.e., the average token length) exhibit weaker correlations with model performance compared to other intrinsic properties.
As compression is a property receiving much focus in work about tokenization,
our findings contribute to the discussion by suggesting that, for character-level tasks, neither the number of tokens nor the degree of compression within a token plays a dominant role in model accuracy.

\subsection{Error Analysis}

\paragraph{Error Overlap between Different Models.}

We compute the intersection-over-union of incorrectly answered questions for each task and pair of models. Figure~\ref{fig:error_overlap_heatmap} presents these ratios for the \texttt{Character Frequency Count} and \texttt{Find Last Occurrence} tasks. For the former, overlap between models is negligible, suggesting that errors do not follow a shared pattern. In contrast, the latter exhibits substantial overlap across most models, indicating more consistent failure cases.
Notably, GPT-4o is an outlier and shows minimal overlap with the other models.

\paragraph{Counting Bias. } We also measure if there exists a systematic bias in the models' predictions. For example, do the models tend to overcount in counting tasks, or over-index in indexing tasks?

We find that there is large variance across models and tasks.
In general,
on the \emph{find\_last\_occurrence} task, models tend to over-index (which can be useful as a guessing bias), whereas
on the \emph{count\_character\_frequency} 
task, models tend to undercount.


\paragraph{Mixed-Case Effects.}

In the evaluation prompt, models are explicitly instructed to treat upper-case and lower-case as different characters for the purpose of the question. We examine whether this aspect of upper/lower case has any effect on model performance.
We find that the presence of mixed case characters appears to introduce a measurable bias, yet no a  catastrophic one. In fact, for some models and tasks, mixed-case inputs seem to improve performance.


\section{Conclusion}

In this work, we introduced a large-scale benchmark for evaluating subword phenomena in character-level tasks. Evaluating tokenization effects in models remains both computationally and architecturally challenging, requiring innovative approaches to conduct such analysis reliably at scale. We hope the benchmark and evaluation measures presented here will support broader efforts to improve model performance on these seemingly simple tasks.

\section*{Limitations}
Our evaluation is limited to character-level tasks in English, chosen to minimize potential confounders such as frequency in the training data. We encourage future work to replicate these experiments in additional languages, prioritizing diversity in typology and script.

\section*{Acknowledgments}
We would like to thank anonymous reviewers for many helpful suggestions.
This research was supported by the Israel Science Foundation (grant No. 1166/23).

\appendix
\section*{Appendix A: CharBench Prompt Template}
\refstepcounter{section}
\label{prompt_template}

We use the following generic prompt format for all CharBench evaluations:

\begin{quote}
    \textit{Answer this question only with the final number, without any other text. Lowercase and uppercase letters are considered different characters.}

    \textbf{Question:} \{question\}
\end{quote}
The placeholder \{question\} is replaced with the exact task instance for each example. The instruction enforces numeric-only answers and case sensitivity, and no additional context is provided beyond this template.

\section{Appendix B: Error Overlap between Different Models}
\refstepcounter{section}
\label{error_overlap}
For each task and model pair, we compute the intersection over union of their error sets, using only items with valid evaluations. Values range from 0 when errors are disjoint to 1 when the error sets coincide.

\begin{figure}[htbp]
    \centering
    \includegraphics[width=\columnwidth]{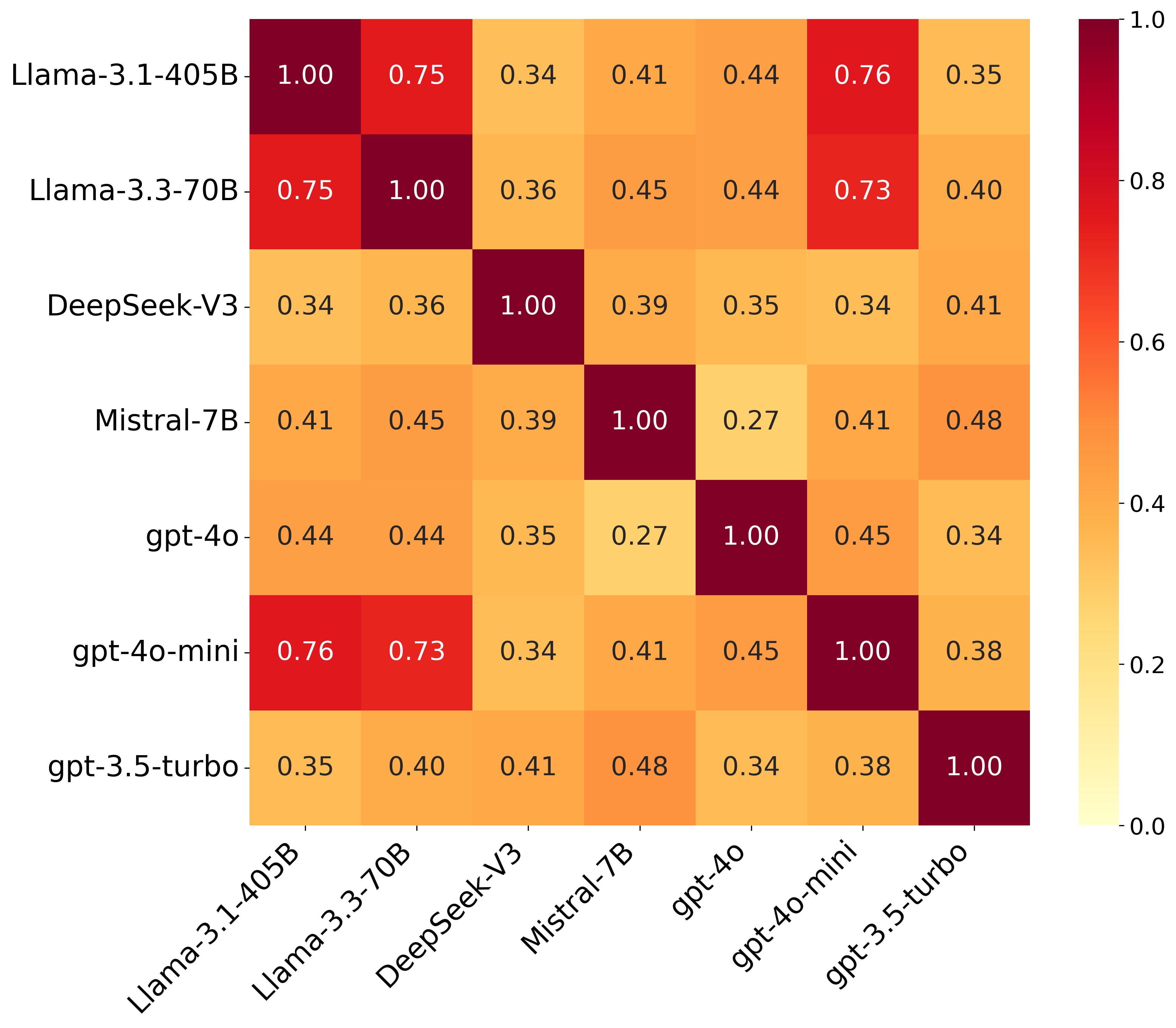}
    \caption{Error overlap for the \emph{count\_unique\_chars} task.}
    \label{fig:heatmap1}
\end{figure}

\begin{figure}[htbp]
    \centering
    \includegraphics[width=\columnwidth]{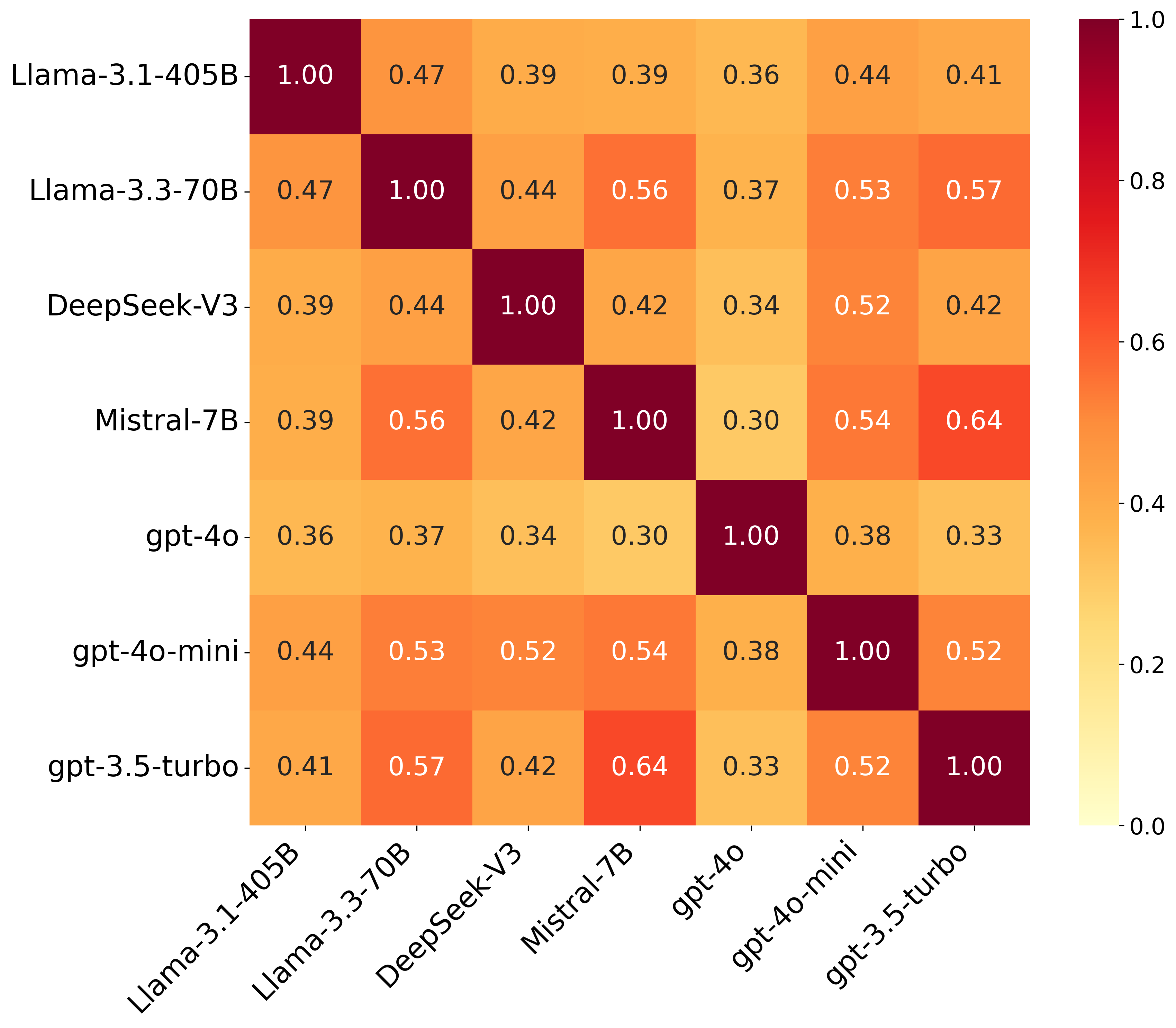}
    \caption{Error overlap for the \emph{find\_last\_occurrence} task.}
    \label{fig:heatmap2}
\end{figure}

\section{Appendix C: Counting Bias}
\refstepcounter{section}
\label{count_bias}

Counting bias measures the average signed error in a model’s numeric predictions for each task. Positive values indicate over counting or over indexing, and negative values indicate under counting or under indexing. We report this per model and task.

\begin{figure}[htbp]
    \label{bias1}
    \centering
    \includegraphics[width=\columnwidth]{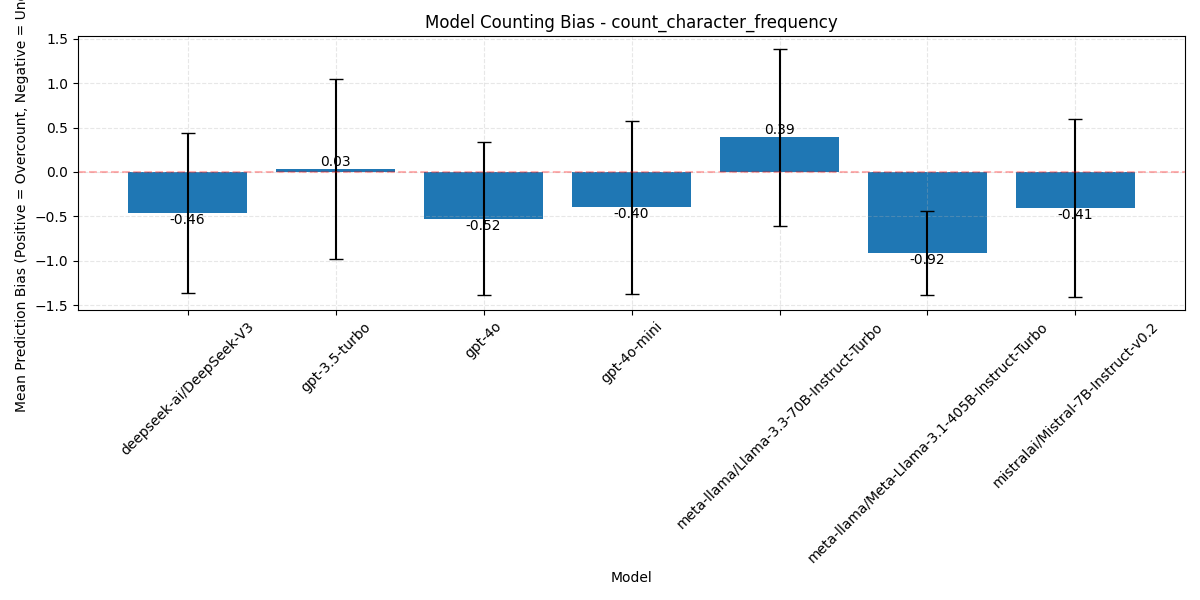}
    \caption{Counting bias across models for the \emph{count\_unique\_chars} task.}
    \label{fig:bias1}
\end{figure}

\begin{figure}[htbp]
    \label{bias2}
    \centering
    \includegraphics[width=\columnwidth]{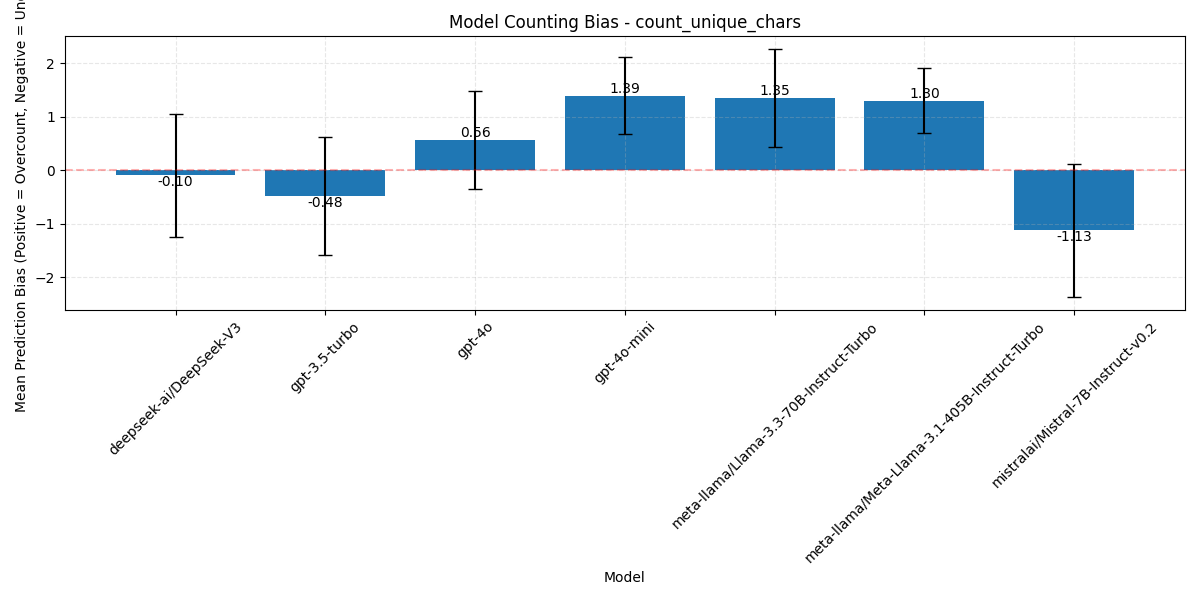}
    \caption{Counting bias across models for the \emph{count\_character\_frequency} task.}
    \label{fig:bias2}
\end{figure}

\begin{figure}[htbp]
    \label{bias3}
    \centering
    \includegraphics[width=\columnwidth]{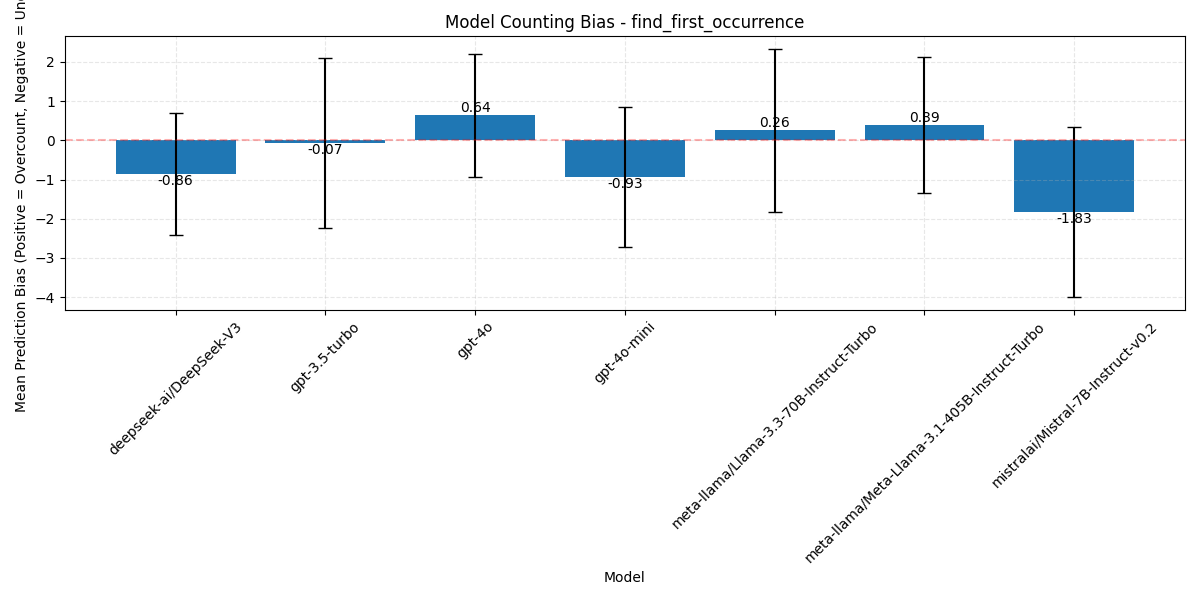}
    \caption{Indexing bias across models for the \emph{find\_first\_occurrence} task.}
    \label{fig:bias3}
\end{figure}

\begin{figure}[htbp]
    \label{bias4}
    \centering
    \includegraphics[width=\columnwidth]{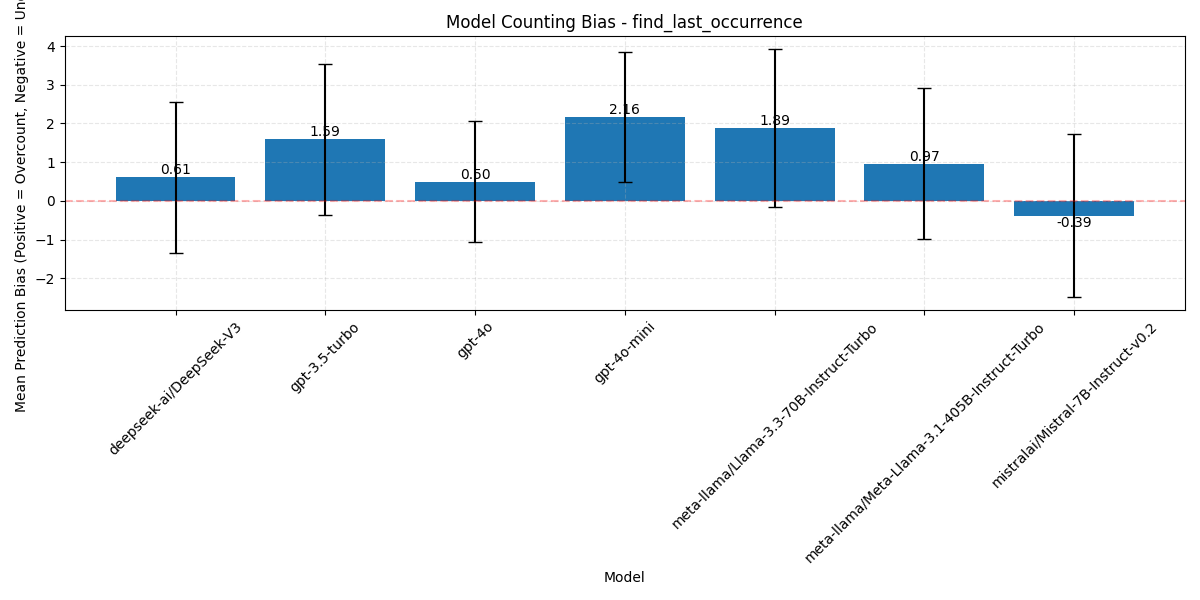}
    \caption{Indexing bias across models for the \emph{find\_last\_occurrence} task.}
    \label{fig:bias4}
\end{figure}

\section{Appendix D: Mixed-Case Effects}
\refstepcounter{section}
\label{mixed_case}
Some items contain both uppercase and lowercase letters. The prompt instructs models to treat case as distinct. We test whether mixed case affects performance: for counting tasks, we flag strings that include both cases; for indexing tasks, we flag items where the target character appears in both its uppercase and lowercase forms.

\begin{figure}[htbp]
    \centering
    \includegraphics[width=\columnwidth]{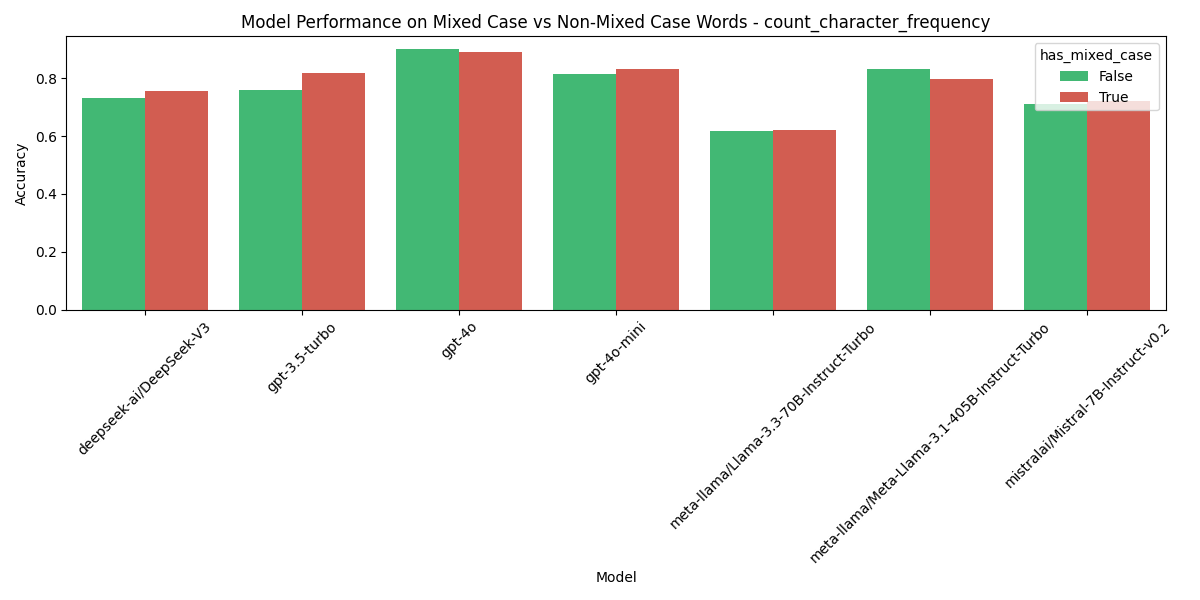}
    \caption{Mixed-case performance for the \emph{count\_unique\_chars} task.}
    \label{fig:mixedcase1}
\end{figure}

\begin{figure}[htbp]
    \centering
    \includegraphics[width=\columnwidth]{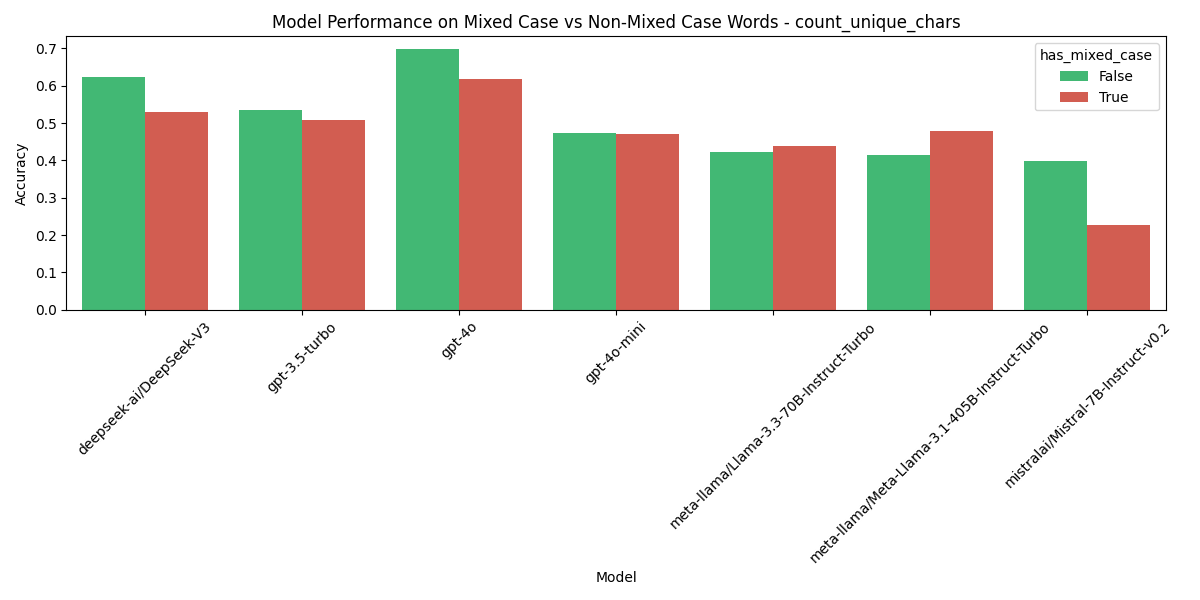}
    \caption{Mixed-case performance for the \emph{count\_character\_frequency} task.}
    \label{fig:mixedcase2}
\end{figure}

\begin{figure}[htbp]
    \centering
    \includegraphics[width=\columnwidth]{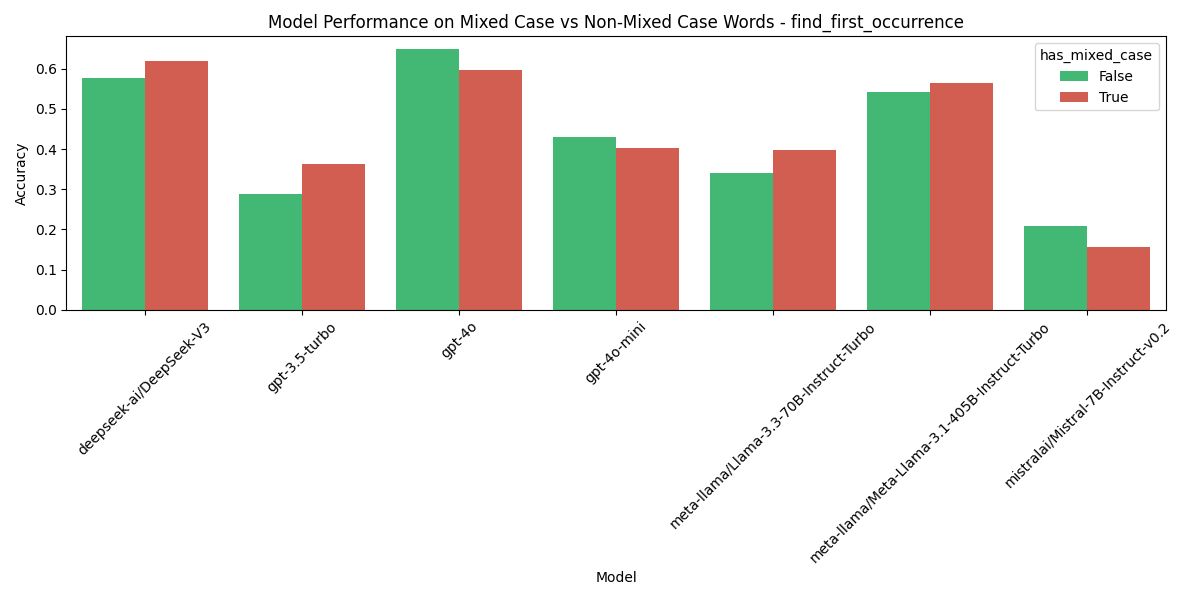}
    \caption{Mixed-case performance for the \emph{find\_first\_occurrence} task.}
    \label{fig:mixedcase3}
\end{figure}

\begin{figure}[htbp]
    \centering
    \includegraphics[width=\columnwidth]{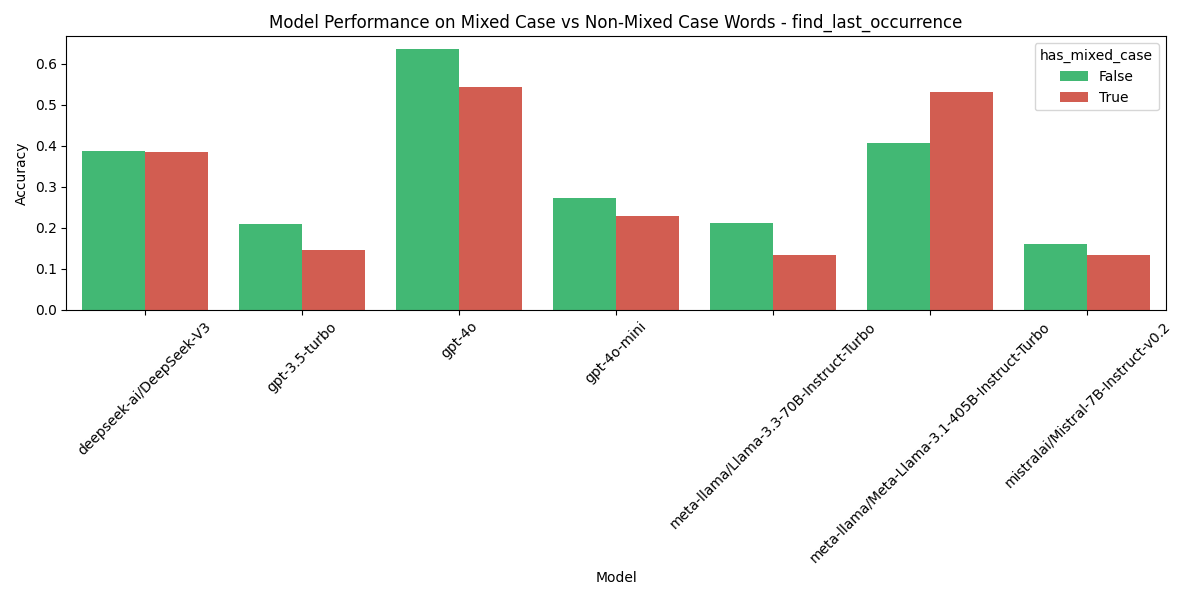}
    \caption{Mixed-case performance for the \emph{find\_last\_occurrence} task.}
    \label{fig:mixedcase4}
\end{figure}

\end{document}